\RequirePackage{amsmath}
\RequirePackage{fix-cm}

\documentclass[twocolumn]{svjour3}          % twocolumn

\smartqed  % flush right qed marks, e.g. at end of proof
\usepackage[pdftex]{graphicx}
\graphicspath{{pdf/}{jpeg/}}
\DeclareGraphicsExtensions{.pdf,.jpeg,.png}

\usepackage{comment}
\usepackage{textcomp}
\usepackage{makecell}
\usepackage{mathtools}
\usepackage{commath}
\usepackage{array,multirow}
\usepackage[normalem]{ulem}
\usepackage{amsmath,amssymb,amsfonts}
\usepackage{algorithmic}
\usepackage{graphicx}
\usepackage{textcomp}
\usepackage{xcolor}
\usepackage{verbatim}
\usepackage[square,numbers]{natbib}
\usepackage{tabularx}

\newcolumntype{Y}{>{\centering\arraybackslash}X}

\begin{document}

\title{Estimating Pose from Pressure Data for Smart Beds with Deep Image-based Pose Estimators}

\titlerunning{Estimating Pose from Pressure Data for Smart Beds with Deep Image-based Pose Estimators}

\author{Vandad Davoodnia \and Saeed Ghorbani  \and Ali Etemad}

\institute{V. Davoodnia \at
              Department of Electrical and Computer Engineering,
              Queen's University, Kingston, Canada \\
              \email{vandad.davoodnia@queensu.ca}           %  \\
           \and
          S. Ghorbani \at
              Department of Electrical Engineering and Computer Science, 
              York University, Toronto, Canada  \\
              \email{saeed@eecs.yorku.ca}           %  \\
            \and
            A. Etemad \at
              Department of Electrical and Computer Engineering,
              Queen's University, Kingston, Canada \\
              \email{ali.etemad@queensu.ca}           %  \\
}

% \date{Received:   \hspace{1cm} / Accepted:  \hspace{1cm} }

\maketitle

\begin{abstract}
In-bed pose estimation has shown value in fields such as hospital patient monitoring, sleep studies, and smart homes. In this paper, we explore different strategies for detecting body pose from highly ambiguous pressure data, with the aid of pre-existing pose estimators. We examine the performance of pre-trained pose estimators by using them either directly or by re-training them on two pressure datasets. We also explore other strategies utilizing a learnable pre-processing domain adaptation step, which transforms the vague pressure maps to a representation closer to the expected input space of common purpose pose estimation modules. Accordingly, we used a fully convolutional network with multiple scales to provide the pose-specific characteristics of the pressure maps to the pre-trained pose estimation module. Our complete analysis of different approaches shows that the combination of learnable pre-processing module along with re-training pre-existing image-based pose estimators on the pressure data is able to overcome issues such as highly vague pressure points to achieve very high pose estimation accuracy.
\end{abstract}

\section{Introduction} \label{sec:sec1}
Sleep studies have recently attracted considerable attention due to the availability and popularization of sensing and processing tools for monitoring users with smart bed technologies. Such technologies play a critical role towards pervasive and unobtrusive sensing and analysis of people in smart homes as well as clinical settings, which in turn can have implications for health, quality of life, and even security. For example, it has been previously demonstrated that different sleeping poses can impact certain conditions or disorders such as sleep apnea \citep{lee2015changes}, pressure ulcers \citep{woo2017exploration}, and even carpal tunnel syndrome \citep{mccabe2011preferred, mccabe2010evaluation}. As another example, in specialized units, the movements of hospitalized patients are monitored to detect critical events and to analyze parameters such as lateralization, movement range, or the occurrence of pathological patterns \citep{cunha2016neurokinect}. Moreover, patients are usually required to maintain specific poses after certain surgeries or procedures to obtain better recovery results. Therefore, long-term in-bed monitoring and automated detection of poses is of critical interest in health-care applications \citep{liu2017vision, lin2017patient}.

Currently, most in-bed examinations are performed with manual visual inspections by caretakers or reports from patients themselves, which are prone to subjective prognosis and user errors. To address the underlying problems in subjective and manual inspections, automated in-bed pose estimation systems are needed in clinical and smart home settings. A number of different learning-based approaches have recently been developed to minimize manual involvement and provide more consistent and accurate results \citep{achilles2016patient}.

Automatic in-bed pose monitoring can be achieved by Deep Neural Networks (DNN),  which provide rich information using convolutional operations for feature extraction on different modalities, such as pressure mapping sensors \citep{davoodnia2020deep, javaid2017balance} or camera-based systems \citep{liu2019seeing}. Camera-based systems, suffer from a range of implementation issues when trying to address challenging situations including blanket occlusions \citep{achilles2016patient}, lighting variations \citep{lLiu2017}, and concerns regarding privacy for in-home and clinical use. Additionally, accurate visual monitoring may require advanced sensors such as infrared \citep{xiao2020infrared, liu2019seeing}, time-of-flight \citep{ruvalcaba2018object}, and depth cameras \citep{he2018pelvic}. The disadvantages of pressure-based systems, on the other hand, are the high cost and need for calibration. Nonetheless, they are not subject to occlusion or point-of-view problems, complications caused by lighting variations, and privacy issues. Moreover, textile-based pressure sensors can be seamlessly embedded into mattresses to construct unobtrusive smart beds \citep{lee2015conductive}.

Recent studies on pressure mapping systems have generally been limited to coarse posture identification (i.e. left, right and supine) \citep{PhysioNet, ostadabbas2014bed, davoodnia2019identity}. Moreover, the notion of body pose estimation using pressure arrays has rarely been explored \citep{liu2014bodypart, davoodnia2019bed, casas2019patient}. The limited scope of existing works on pressure data is mainly due to the lack of extensive datasets that span the pose and body distributions required to learn generalized models for pose estimation.

\begin{figure}
\begin{center}
{\includegraphics[width=0.95\linewidth]{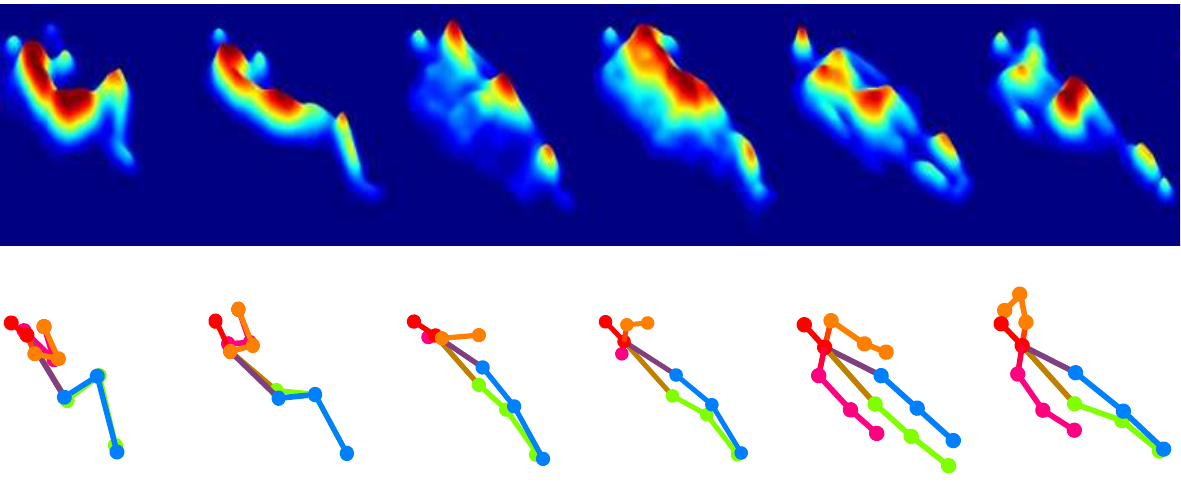}}
\end{center}
   \caption{Some examples of pressure-based in-bed pose estimation are presented. The top row shows the input pressure maps recorded using a mattress with embedded sensors, while the bottom row presents the estimated poses using an end-to-end method consisting of our proposed PolishNetU network with OpenPose.}
\label{fig:teaser}
\end{figure}

With recent advances in machine and deep learning, a large number of advanced data-driven pose estimation methods have been developed \citep{tang2018deeply, ke2018multi, yang2017learning, chen2017adversarial, chu2017multi, cao2016realtime, insafutdinov2016eccv, insafutdinov2017cvpr, chen2018cascaded} to be used in a variety of different applications such as animation, clinical monitoring, human-computer interaction, and robotics \citep{koppula2016anticipating, tulyakov2017mocogan, imabuchi2018automated}. While these models have the potential to be used for pressure-based in-bed pose estimation, they have all been designed and trained for \textit{images} with naturally appearing human figures, and mostly in upright postures. Moreover, \textit{weak pressure areas} resulting from supported body parts are perceived as body occlusions in the pressure maps, which can cause low fidelity with existing models.

In this paper, we explore the use of pre-existing \textit{image-based pose estimators}, namely OpenPose \citep{cao2016realtime} and Cascaded Pyramid Network (CPN) \citep{chen2018cascaded}, for in-bed pressure-based pose estimation, with the goal of detecting keypoint locations reliably in challenging conditions such as weak pressure areas. See Figure \ref{fig:teaser} (top row) for a few examples of the pressure data used in this study. To this end, we exploit four general approaches: 
\textit{i}) First, as a baseline, we use the pose estimators pre-trained on their respective image datasets without any modification or training with the pressure data; 
\textit{ii}) Next, we manually label body joint locations for the pressure dataset and re-train the pose estimators from the previous pipeline on pressure datasets. 
\textit{iii}) We propose a fully-convolutional network called PolishNetU, to learn a pre-processing step such that the polished outputs are consistent with the data on which the pose estimators have been originally trained. PolishNetU is then followed by the frozen pose estimators (not re-trained or modified for pressure data).
\textit{iv}) Finally, we re-train the entire pipeline \textit{iii} (consisting of PolishNetU and a pose estimator) on the labeled pressure data. Our analysis shows that method (\textit{ii}) and (\textit{iv}) perform superior compared to other approaches, followed by solution (\textit{iii}), which achieves significantly better results compared to the mere use of a pose recognition model directly on the inputs (pipeline (\textit{i})). In addition, we show that while most pose estimation models require large datasets for training, PolishNetU is capable of generalizing when trained on a dataset with a limited number of subjects. In summary, our contributions are as follows:
\begin{itemize}
    \item We explore the notion of using existing pose recognition networks for in-bed pose estimation using pressure data and improve upon the previous studies.
    \item We propose PolishNetU, a deep UNet-style neural network for taking pressure images and transforming them into an embedding close to real human figures captured in images.
    \item We compare and provide comprehensive insights into several strategies for using pre-existing pose estimators namely OpenPose and CPN for pressure-based pose estimation.
    \item Finally, we conclude that fine-tuning pre-existing pose estimators along with using PolishNetU as a pre-processing domain adaptation step can perform pose recognition very effectively with a detection rate of over $99\%$.
\end{itemize}

\section{Related Work} \label{sec:sec2}
Generally, in-bed pose estimation methods can be divided into two main categories based on the input modality: camera-based and pressure-based. The former are a group of techniques that make use of different types of cameras such as infrared, range, or normal digital cameras, while the latter use a matrix of pressure sensors. While our work focuses on pressure data, we review both approaches in this section for completeness and providing a complete picture of solutions used for in-bed pose estimation.

\subsection{Camera-based Pose Estimation}
These methods suffer from occlusion and lighting variation conditions. As a result, one of the main focuses of these approaches is to address such problems. For example, \citep{lLiu2017} proposed a novel recording method, called Infrared Selective (IRS) image acquisition, to address the problem of lighting variations caused by the daylight cycle. Then an $n$-end Histogram of Oriented Gradients (HOG) feature extraction followed by a Support Vector Machine (SVM) classifier was used to align the orientation such that the rectified images were consistent with common pose detection methods. Finally, a few layers of a convolutional pose machine were fine-tuned on the in-bed dataset. Their pose rectification and estimation blocks assume no occlusion, for example by a blanket, and use an on-demand trigger to reduce the high computational cost of the pipeline. In another camera-based work, \citep{Wang2010} developed a video-based monitoring approach to estimate human pose in conditions with occluded body parts. The proposed method comprised two main blocks: a weak human model and a modified pose matching algorithm. First, in order to reduce the search space of poses, a weak human model was used to quickly generate soft estimates of obscured upper body parts. The obscured parts were then detected using edge information in multiple stages. Next, an enhanced human pose matching algorithm was introduced to address the problem of weak image features and obstruction noise. This was used as a subsequent fine-tuned block to be optimized in the constrained space.

End-to-end deep learning methods have also been explored for in-bed camera-based pose estimation. Achilles et al. \citep{achilles2016patient} trained a deep model to infer body pose from RGB-D data, while the ground truth was provided by a synchronized optical motion capture system. The model was constructed by a convolutional neural network (CNN) followed by a recurrent neural network (RNN) to capture the temporal consistency. Since it was impossible to track the markers while occluded by a blanket, the RGB-D data were augmented with a virtual blanket to simulate the conditions where body parts were occluded. 

In \citep{Chen2018}, a semi-automatic approach was proposed for upper-body pose estimation using RGB video data. The video data were normalized in a pre-processing step using contrast-limited adaptive histogram equalization, making the processed data invariant to lighting variations. Then, a CNN model was trained on the subsequent data for each subject outputting $7$ heatmaps for $7$ upper-body joint locations. Finally, a Kalman filter was applied as a post-processing step to refine the predicted joint trajectories and achieve a more temporally consistent estimation.

\subsection{Pressure-based Pose Estimation}
Pressure-based approaches have recently attracted attention as they avoid some of the problems that camera-based systems suffer from, for example, occlusion, lighting variations, and subject privacy. In \citep{ostadabbas2014bed}, subject classification was performed with pressure data in three standard postures, namely supine, right side, and left side. Eighteen statistical features were extracted from the pressure distributions in each frame of each posture and fed to a dense network. Hidden layers were pre-trained by incorporating restricted Boltzmann machines into the deep belief network to find the proper initial weights.

In \citep{Grimm2011}, a generative inference approach was proposed similar to \citep{Singh2017}. However, pressure data were used as the input modality, and the body was simulated using a less sophisticated human body model. The pipeline included two main blocks. First, the patient orientation was detected and then the coarse body posture was classified using a $k$-nearest-neighbor classifier by comparing the query pressure distribution to the labeled training data. In the second step, a cylindrical $3$D human body model was used in a generative inference approach to synthesize pressure distributions. The body model parameters (shape and pose) were iteratively optimized using Powell’s method, minimizing the sum of squared distances between the synthetic pressure distribution and observed distribution. 

In a more recent study, \citep{clever2020bodies} proposed PressureNet, a pressure-based $3$D pose and shape reconstruction network, which was trained on synthetic data and tested on real pressure images. Their method consisted of two modules, first, to encode shape, pose, and global transformation from the gender and pressure data, and second, for reconstructing the $3$D model and consequently estimating the pressure images from first module's input pose information. By incorporating pose information loss for the first module and heatmap loss for the second, they were able to achieve a $3$D pose recognition error of less than $75mm$.

\begin{figure*}
\begin{center}
{\includegraphics[width=0.9\textwidth]{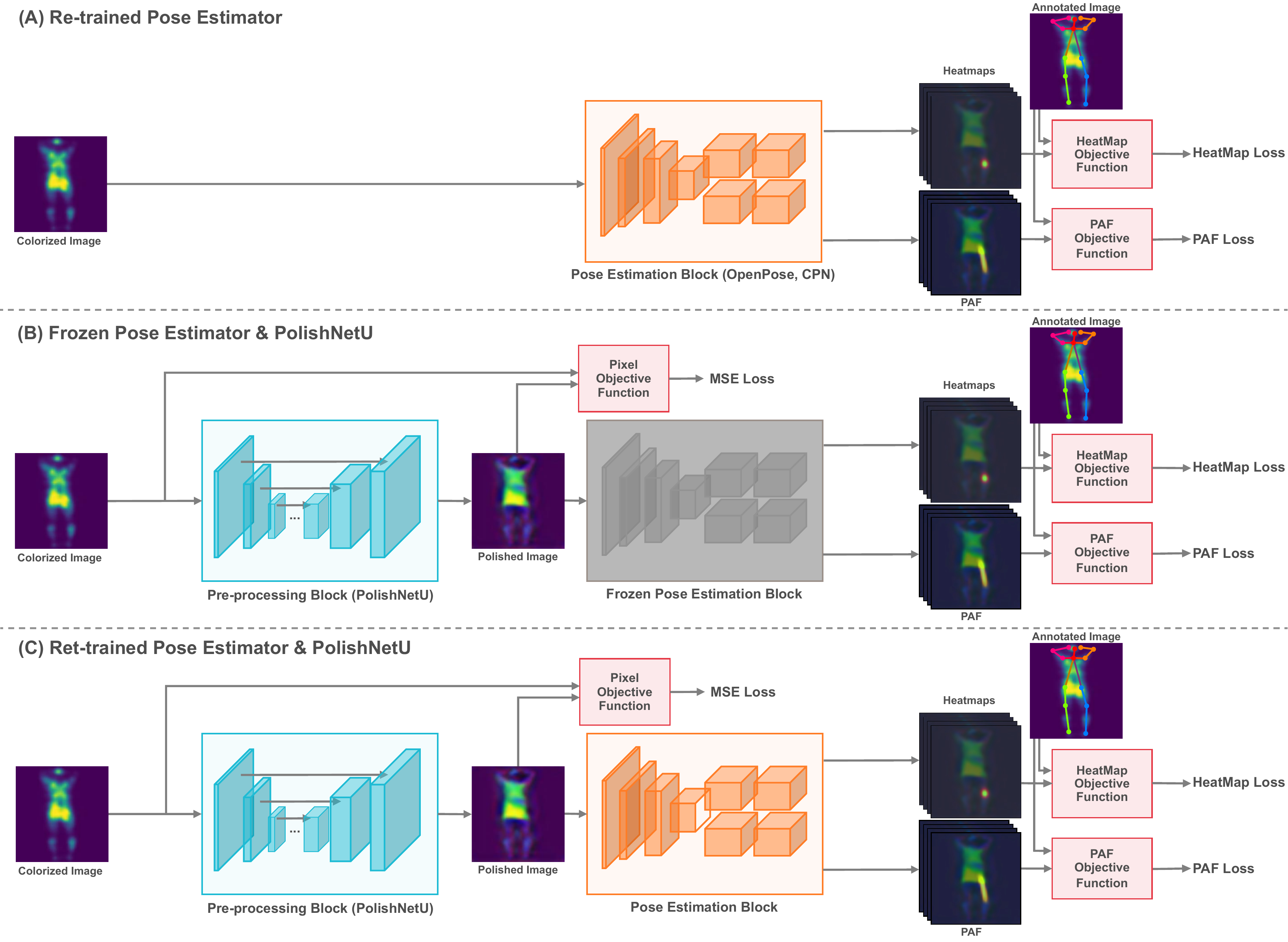}}
\end{center}
   \caption{The explored methods for in-bed pose estimations and respective loss functions are presented: (A) The off-the-shelf pose estimation network is used either as is or trained on the pressure data and annotated keypoints. (B) PolishNetU is utilized to pre-process the input pressure data and generate outputs close to realistic images to feed into the pre-existing pose estimation networks. (C) Similar to previous approach, but the pose estimation network is re-trained, either alone or together with PolishNetU.}
\label{fig:network_architecture}
\end{figure*}

\section{Methodology}
\subsection{Problem Setup}
Our goal is to explore possible approaches for in-bed pose estimation using pre-existing pose recognition networks. Our problem can be formulated as desiring a set of $14$ keypoints, indicating different limb positions, by taking an input pressure image and passing it through a deep neural network. In this paper, we analyze a set of different solutions where pose estimation is achieved by utilizing off-the-shelf pose estimators. To this end, the explored networks may include a learnable pre-processing network (PolishNetU), which aims to edit the pressure images to prepare them for pose identification. In the PolishNetU model, the latent features are learned through the multi-scale architecture of the U-Net style network. On the encoder part of PolishNetU, the latent features of each step are pooled with the previous blocks to be used to generate a polished and pre-processed image in the decoder part. The polished image is then passed to a pose estimation block, which locates the position of different joints. As mentioned in Section 1 (Introduction), we consider OpenPose and CPN for pose estimation. These methods are from a family of convolutional pose machines \citep{wei2016convolutional}, where they use a VGG and a ResNet backbone as feature extractors respectively, and their goal is to find the keypoints by estimating the probability of the existence of the limb in the image. As a result, both CPN and OpenPose generate $14$ heatmaps, each corresponding to one limb position. Furthermore, OpenPose incorporates additional $28$ output channels called Part Affinity Fields (PAF) corresponding to the connection of the adjacent limbs and their difference in position. See Figure \ref{fig:network_architecture} for an overview of the explored approaches.

\subsection{Solutions} \label{sec:solutions}
We consider $4$ possible solutions: (\textit{i}) first, the off-the-shelf pose estimators can be used without any further training or addition of additional modules; (\textit{ii}) the pose estimation networks can be re-trained on the pressure data; (\textit{iii}) a domain adaptation network can be designed and trained to pre-process the vague input pressure maps to then be used with the frozen pose estimators; and (\textit{iv}) the entire pipeline consisting of the domain adaptation network and the pose estimator can be re-trained end-to-end with the pressure data. Following we describe the details of each of these possible solutions.

\textbf{Frozen pose estimators} utilized directly on the pressure images. Let $I \in \mathbb{R}^{W \times H \times 3}$ be the unpolished input pressure data. Our objective would then be to estimate a set of keypoints $\hat{K} = Q(I; \theta_Q)$, where $Q$ is the approximated function by the pre-existing pose recognition networks and $\theta_Q$ is its pre-trained parameters. In this scenario, the pre-trained parameters $\theta_Q$ are optimized on the original image-pose dataset that the pose estimator has been trained on. 
 
\textbf{Re-training Pose Estimators} is the second possible solution which is built upon the previous one by adding the objective of optimizing $\theta_Q$ on the pressure data such that $Q$ closely estimates the keypoints $K$. We define the objective as:
\begin{equation}
% \begin{aligned}
~~~~~~~~~~~~~~~~~~~~~~~~~~\max_{\theta_P} J(\hat{K},K) ,
% \end{aligned}
\end{equation}
where $J$ is defined as a function of similarity between the predicted and ground truth keypoints. Accordingly, We define two loss functions with a heatmap term and a PAF term. The heatmap term, $E_{heatmap}$ is defined as:
\begin{equation}
\begin{aligned}
E_{heatmap} = \frac{1}{KHW}\sum_{k=1}^{K}{\sum_{i=1}^{H}{\sum_{j=1}^{W}{V_k\left \| C_k - C_k' \right \|_{2}^{2}}}} \,\,\,,
\end{aligned}
\end{equation}
where $C_k$ and $C_k'$ are the corresponding ground-truth and predicted heatmaps for keypoint $k$, $K = 14$ is the number of visible keypoints, and $V_k$ is $1$ if the $K_{th}$ limb is visible and $0$ if it's not. Next, the PAF term, $E_{PAF}$, is defined as:
\begin{equation}
\begin{aligned}
E_{PAF} = \frac{1}{LHW}\sum_{l=1}^{L}{\sum_{i=1}^{H}{\sum_{j=1}^{W}{V_l\left \| F_l - F_l' \right \|_{2}^{2}}}} \,\,\,,
\end{aligned}
\end{equation}
where $F_l$ and $F_l'$ are the corresponding ground-truth and predicted PAFs for limb $l$, $L = 28$ is the number of connections between the limbs in $y$ or $x$ axis, and $V_l$ is $1$ if both limbs producing the $l_{th}$ connection are visible and $0$ otherwise. The final objective would then be optimized by minimizing the sum of the two loss functions.

\textbf{Image space representation learning} is the next possible solution. Here, our goal is to implement a learnable pre-processing step that receives the pressure data $I$ as inputs and synthesizes colored images close to a pre-trained pose estimation network's learnt data. Therefore, the output data from the learner should lie on the data manifold by which the pose estimation module was trained. This learnable pre-processing step, which we call PolishNetU, converts the pressure data to polished images that better resemble human figures as expected by common pose estimation models.

Lets define $I' \in \mathbb{R}^{W \times H \times 3}$ as the output of our PolishNetU $\textit{P}$, in other words $I' = \textit{P}(I; \theta_{\textit{P}})$ and $\hat{K} = Q(I'; \theta_Q)$, where P is the pre-processing function, and $\theta_{\textit{P}}$ is its set of trainable parameters. Using the $E_{heatmap}$ and $E_{PAF}$, a pre-trained pose estimation module can force PolishNetU to synthesize entirely new images. To prevent pose deviations, we also added a third term, pixel loss, to the objective function. This term acts as a regularizer, which penalizes the distance between the input pressure maps and the synthesized polished images, which we defined as:
\begin{equation}
\begin{aligned}
E_{pixel} = \frac{1}{HW}\sum_{i=1}^{H}{\sum_{j=1}^{W}{\left \| I - I' \right \|_{2}^{2}}} \,\,\,.
\end{aligned}
\end{equation}

Finally, we optimize $\theta_{\textit{P}}$ using the objective function:
\begin{equation}
\begin{aligned}
E(\theta_{\textit{P}}) = \lambda_{heatmap}E_{heatmap} & +\lambda_{PAF}E_{PAF} \\ 
& +\lambda_{pixel}E_{pixel}  \,\,\, ,
\end{aligned}
\end{equation}
where we chose $\lambda_{heatmap}=\lambda_{PAF}$ and adjusted $\lambda_{pixel}$ to achieve the optimum representations. It is important to note that the pixel loss weight $\lambda_{pixel}$ has a large impact on the representations, where large values can result in PolishNetU producing outputs overly similar to the unpolished inputs. In contrast, smaller values caused the heatmap and PAF losses to overlook this regularizer, resulting in significant deviations from the input images, generating non-human like images. Therefore in this setup, we first train the network by setting a high value for $\lambda_{pixel}$, and then slowly decrease its weight to enable our networks to focus on reconstructing weak pressure points after stabilizing.

\textbf{End-to-end re-training of PolishNetU and pose estimators} are the final possible solution. Since PolishNetU is acting as the pre-processing step, it is intuitive that the main pose estimation network would have a large impact on the performance. Therefore, after learning the real image representations $I'$ by training the PolishNetU, we move on to optimizing $\theta_Q$ by using the same objective function. Additionally, we can also choose to fine-tune PolishNetU alongside our pose estimation network, which we refer to as fine-tuned PolishNetU in the future sections.

\section{Experiments and Results}
\subsection{Data Preparation}
We use two different datasets to train and test our pressure-based pose estimation approach, PmatData \citep{ostadabbas2014bed} available in the PhysioNet repository \citep{goldberger2000physiobank} and HRL-ROS dataset \citep{clever20183d}. These datasets are briefly described as follows:

\textbf{PmatData} \citep{ostadabbas2014bed} was recorded by a force sensitive pressure mapping mattresses. Each mattress contained $2048$ sensors spread on a $32$ by $64$ grid with each sensor being $25.4$ \textit{mm} apart. The recording was performed with a frequency of $1$ Hz for a pressure range of $0-100$ mmHg. Data were recorded from $13$ healthy subjects in $8$ standard postures and $9$ further sub-postures, for a total of $17$ unique pose classes. Subjects were within a height range of $169-186$ \textit{cm}, a weight range of $63-100$ \textit{Kg}, and an age range of $19-34$ years. We developed and utilized a tool in MATLAB for annotating the body part keypoints in $18256$ data samples. The annotating procedure was carried out by two researchers and then cross-checked to ensure consistency. To perform very rigorous evaluation experiments, we employ a leave-some-subjects-out validation strategy, training the network on $9$ subjects and testing it on the remaining $4$.

\textbf{HRL-ROS} \citep{clever20183d} was collected for kinematic-based $3$D pose recognition using a configurable bed embedded with pressure sensors and motion capture cameras. The bed was equipped with an array of $27\times64$ sensors distributed $28.6$ \textit{mm} apart. A total of $17$ subjects were asked to lie or sit in different postures and move a body limb in a specific path, while their limb position was being tracked using motion capture cameras. Subjects were within a height range of $160-185$ \textit{cm}, a weight range of $45.8-94.3$ \textit{Kg}, and an age range of $19-32$ years. For our purposes we use the data from all of the subjects in all $13$ lying postures, resulting in a total of $39095$ pressure maps. Similar to PmatData, we leave the last $4$ subjects for testing while keeping the rest for training.

\subsection{Pre-processing}
First, we remove the noise caused by occasional malfunctioning pressure sensors. These artifacts usually occur when certain individual sensors become subject to pressure values outside the calibrated voltage range. To clean up the pressure values, we use a $3 \times 3 \times 3$ spatio-temporal median filter. We tune the filter size by evaluating the pose estimation performance of a frozen OpenPose on the PmatData dataset, showing that larger filter sizes do not improve the performance. Following previous studies \citep{davoodnia2019bed, davoodnia2019identity, davoodnia2020deep}, we also remove the first $3$ frames of each sequence, which in some cases are transition frames where pressure maps are not clear. We identified the frames and cross-checked them visually to remove the outliers using the histogram of the dataset based on the average pressure of each image.

\begin{figure*}
\begin{center}
{\includegraphics[width=\textwidth]{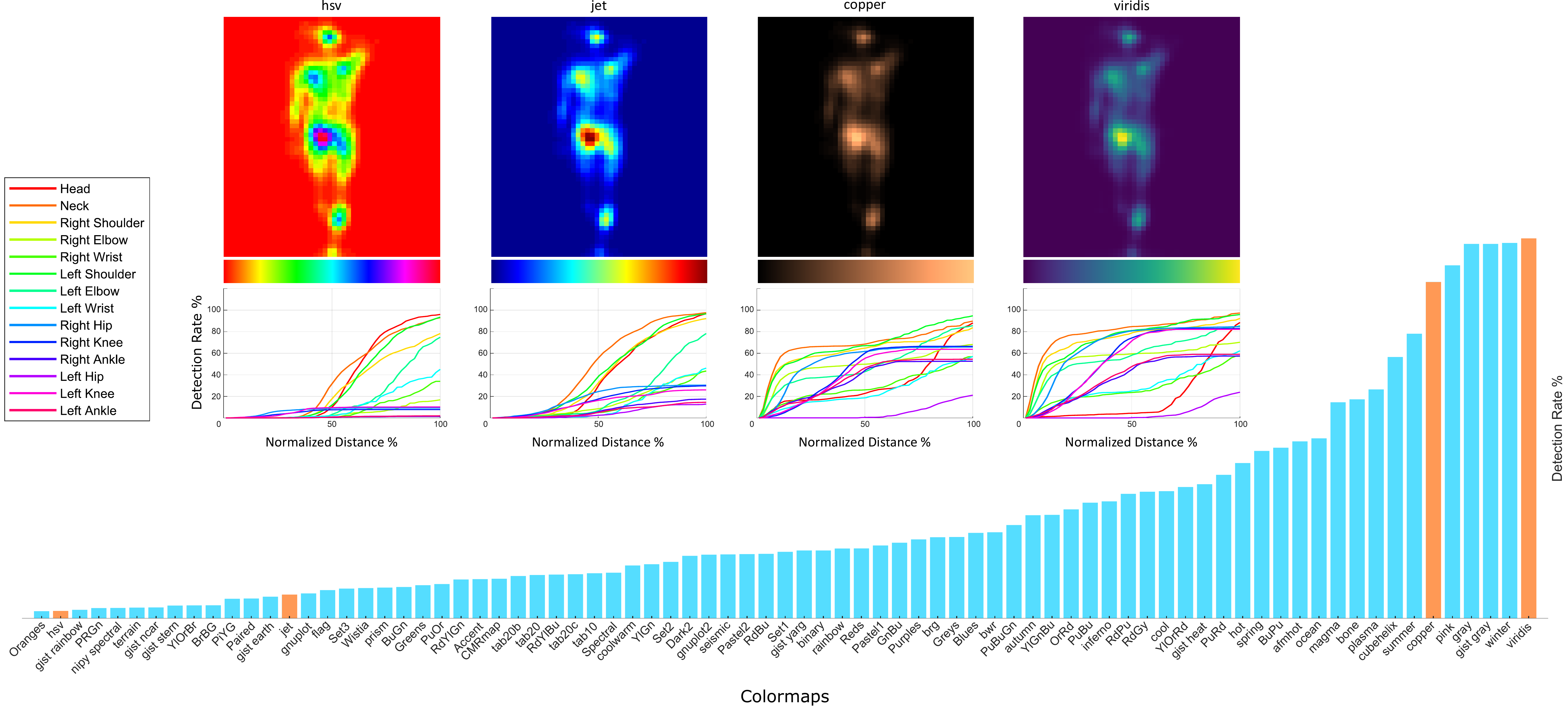}}
\end{center}
   \caption{We compared the effect of applying $38$ different colormaps to the pressure data by benchmarking the performance of OpenPose on the colorized images. Subsequently, the Viridis colormap achieved the highest average detection rate among all the available schemes.}
\label{fig:color_map}
\end{figure*}

The input pressure maps are provided in the form of $W \times H \times 1$ arrays. However, most existing pose estimation methods have been trained on color images. As a result in order to utilize existing frameworks for our goal of estimating poses from pressure data, the pressure maps need to be converted to color images. Consequently, we convert the pressure maps to color images using a colormap. To this end, we need to select a colormap with high compatibility with the pose estimation network, and in general, with natural images containing human figures. Our investigations show that the choice of colormap can play a considerable role in the performance. We investigate $38$ different colormaps and evaluate the error rates when PolishNetU is excluded from the pipeline and only the frozen OpenPose model is used. Figure \ref{fig:color_map} shows the performance over different colormaps and illustrates the estimated pose accuracy for different body parts for $4$ sample colormaps from the distribution, namely HSV, Jet, Copper, and Viridis. Eventually, as shown in Figure \ref{fig:color_map}, Viridis exhibits the best color mapping characteristics. The pose estimation evaluation method is presented in Section \ref{evaluation}.

\subsection{Implementation Details}
\subsubsection{PolishNetU}
Polishing the colorized pressure maps and converting them to images consistent with the pose estimation module is done by a feed-forward network $\textit{P}$ called PolishNetU. Our proposed network is a variation of U-Net, consisting of a combination of encoders and decoders, including fully convolutional and deconvolutional layers, respectively. Given the colorized pressure data, a series of $8$ encoder blocks of Conv-BatchNorm-LeakyReLu with a stride of $2$ are exploited to encode the input to a data manifold, capturing pressure properties that incorporate pose information. The polished image space is then achieved by concatenating the residuals from the encoders with UpSample-DeConv-BatchNorm-LeakyReLu on the encoded latent space. Finally, using a $tanh$ activation function, the last layer provides a polished image, which is compatible with the pose estimation module conditioned on the given pressure input. We use BatchNorm for faster and more reliable training, and upsampling layers instead of a deconvolution layer with a stride of $2$ to avoid the deconvolution checkerboard artifact.

\subsubsection{Pose Estimation Module}
To train PolishNetU, we utilize OpenPose \citep{cao2016realtime} or CPN \citep{chen2018cascaded} as our pose identification module. OpenPose is a well-known network developed for real-time pose estimation first, by which has been recently extended to support face landmark detection and hand gesture detection as well. On the other hand, CPN is a more recent and powerful pose recognition method which was able to achieve state-of-the-art in many of the pose recognition challenges. The authors of OpenPose defined heatmap and Part Affinity Fields (PAF) outputs which we utilize to define our pose estimation objectives. Each heatmap provides a $2$D distribution of the belief that a keypoint is located on each pixel. Additionally, PAF is defined as a $2$D vector field for each limb, where each $2$D vector encodes both position and the orientation of the limb. Over the past few years, several versions of this network have been published, mostly with changes in the final blocks for face and hand landmark detection and trade-offs between memory usage, performance, and speed. For our work, we choose the original version published in $2016$ that includes $7$ stages for refining the heatmaps and PAFs. We use the exact design parameters of the original OpenPose model. In our pipeline, neither OpenPose, nor CPN, do not use any non-linear activation functions, therefore vanishing gradients is not an issue. Furthermore, we neglect the loss for non-visible parts, eyes, and ears due to the nature of our pressure data. Consequently, we end up with $14$ heatmaps for the head, neck, shoulders, elbows, wrists, ankles, knees, and the hip, as well as $28$ PAFs connecting the body parts to be used in our cost functions.

\begin{figure*}[t]
\begin{center}
{\includegraphics[width=\linewidth]{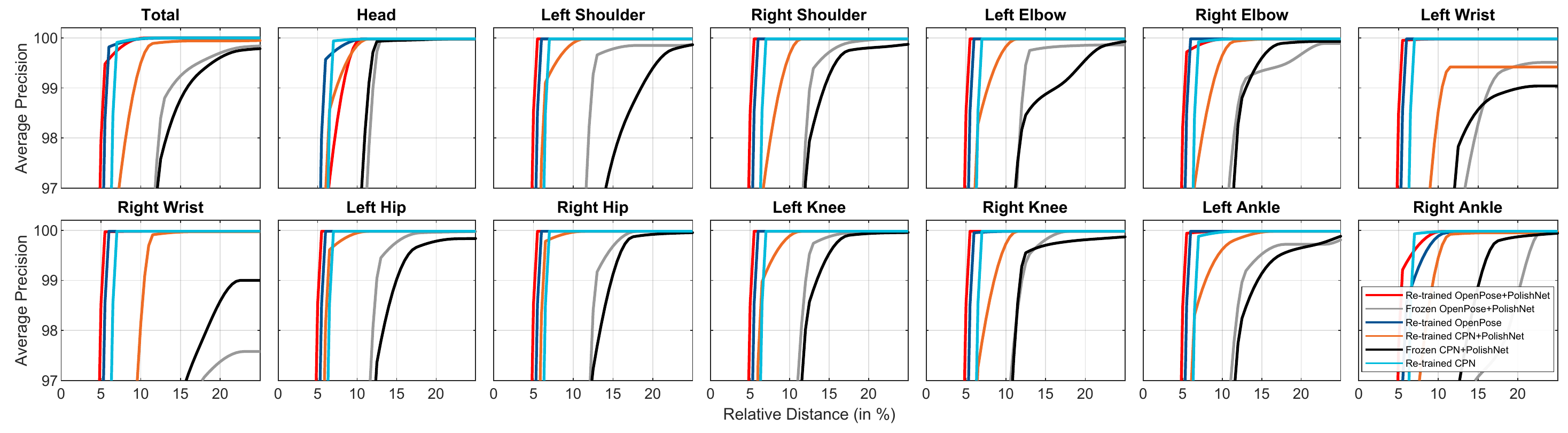}}
\end{center}
   \caption{AP over different thresholds is demonstrated for each body part from PmatData dataset.}
\label{fig:PCK_pmat}
\end{figure*}

\begin{figure*}[t]
\begin{center}
{\includegraphics[width=\linewidth]{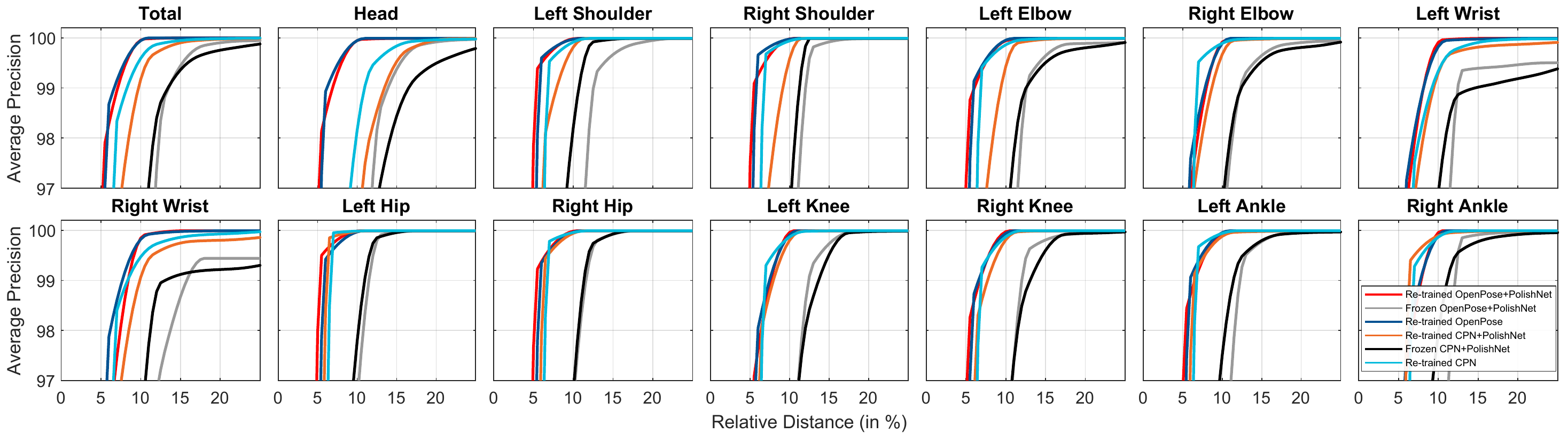}}
\end{center}
   \caption{AP over different thresholds is demonstrated for each body part from HRL-ROS dataset.}
\label{fig:PCK_hrl}
\end{figure*}

\subsubsection{Pipeline}
The pipeline is implemented using TensorFlow on an NVIDIA Titan XP GPU. We use an Adam optimizer at the training stage with a learning rate of $10^{-3}$, which is decayed with a rate of $0.95$ for every $1000$ update iterations. The pipeline is trained for $40$ epochs with a batch size of $16$. $\lambda_{PAF}$ and $\lambda_{heatmap}$ are both set to $1$ while $\lambda_{pixel}$ is changed from $1$ to $0.01$ during training to first stabilize the model, and then allow it to interpolate the vivid body limbs without pixel loss penalty.

\subsection{Performance Evaluation} \label{evaluation}
After obtaining the output heatmaps in the form of $W_h \times H_h \times 14$, we smooth them using a Gaussian kernel of $3 \times 3$ across the spatial dimensions to reduce prediction noise in the output. Then, we perform a flip-test to reduce the model's bias to left and right directions by obtaining the output from the original input and its flipped version and then averaging them. Finally, we take the location of the maximum of each channel as our predicted keypoints, obtaining a $14 \times 3$ array containing the location and prediction scores of the body limbs.

\newcommand{\mcs}[1] {\scriptsize#1}
\newcommand{\mc}[1] {\scriptsize#1}
\begin{table*}
\begin{center}
\caption{The AP5 values are presented for different methods on PmatData.}
\label{table:average_detection_pmat}
\begin{tabularx}{\textwidth}{lYYYYYYYYYYYYY}
\Xhline{2\arrayrulewidth}
\mcs{}                 & \mcs{\textbf{H}}   & \mcs{\textbf{LS}}  & \mcs{\textbf{RS}}  & \mcs{\textbf{LE}} & \mcs{\textbf{RE}}   & \mcs{\textbf{LW}}  & \mcs{\textbf{RW}} & \mcs{\textbf{LH}}   & \mcs{\textbf{RH}}  & \mcs{\textbf{LK}}  & \mcs{\textbf{RK}} & \mcs{\textbf{LA}}   & \mcs{\textbf{RA}}          \\  
\hline
\mcs{{Re-trained OpenPose+PolishNetU}}	&	\mc{99.4}	&	\mc{100.0}	&	\mc{100.0}	&	\mc{100.0}	&	\mc{100.0}	&	\mc{100.0}	&	\mc{100.0}	&	\mc{100.0}	&	\mc{100.0}	&	\mc{100.0}	&	\mc{100.0}	&	\mc{100.0}	&	\mc{99.8}   \\
\mcs{{Re-trained OpenPose}}	            &	\mc{91.3}	&	\mc{100.0}	&	\mc{100.0}	&	\mc{100.0}	&	\mc{100.0}	&	\mc{100.0}	&	\mc{100.0}	&	\mc{100.0}	&	\mc{100.0}	&	\mc{100.0}	&	\mc{100.0}	&	\mc{100.0}	&	\mc{100.0}  \\
\mcs{{Frozen OpenPose+PolishNetU}}   	&	\mc{92.2}	&	\mc{90.1}	&	\mc{45.7}	&	\mc{80.8}	&	\mc{88.5}	&	\mc{52.5}	&	\mc{43.2}	&	\mc{48.1}	&	\mc{36.9}	&	\mc{80.2}	&	\mc{98.6}	&	\mc{100.0}	&	\mc{96.0}   \\
\mcs{{Frozen OpenPose}}	                &	\mc{0.0}	&	\mc{0.0}	&	\mc{0.0}	&	\mc{0.0}	&	\mc{0.0}	&	\mc{0.0}	&	\mc{0.0}	&	\mc{0.0}	&	\mc{0.0}	&	\mc{0.0}	&	\mc{0.0}	&	\mc{0.0}	&	\mc{0.0}    \\
\hline
\mcs{{Re-trained CPN+PolishNetU}}    	&	\mc{99.6}	&	\mc{99.5}	&	\mc{99.5}	&	\mc{99.5}	&	\mc{99.6}	&	\mc{99.4}	&	\mc{99.4}	&	\mc{99.6}	&	\mc{99.5}	&	\mc{99.6}	&	\mc{99.6}	&	\mc{99.5}	&	\mc{99.5}  \\
\mcs{{Re-trained CPN}}	                &	\mc{98.7}	&	\mc{99.3}	&	\mc{99.4}	&	\mc{99.3}	&	\mc{99.3}	&	\mc{98.9}	&	\mc{99.1}	&	\mc{99.5}	&	\mc{99.5}	&	\mc{99.4}	&	\mc{99.3}	&	\mc{99.1}	&	\mc{98.8}  \\
\mcs{{Frozen CPN+PolishNetU}}        	&	\mc{100.0}	&	\mc{72.3}	&	\mc{44.3}	&	\mc{82.0}	&	\mc{78.5}	&	\mc{41.6}	&	\mc{14.4}	&	\mc{48.6}	&	\mc{67.7}	&	\mc{88.9}	&	\mc{100.0}	&	\mc{82.9}	&	\mc{85.4}  \\
\mcs{{Frozen CPN}}	                    &	\mc{0.0}	&	\mc{0.0}	&	\mc{0.0}	&	\mc{0.0}	&	\mc{3.3}	&	\mc{4.7}	&	\mc{0.0}	&	\mc{0.0}	&	\mc{0.0}	&	\mc{7.9}	&	\mc{31.3}	&	\mc{23.1}	&	\mc{4.4}   \\
\Xhline{2\arrayrulewidth}
\end{tabularx}
\end{center}
\end{table*}

\begin{table*}
\begin{center}
\caption{The AP5 values are presented for different methods on HRL-ROS.}
\label{table:average_detection_hrl}
\begin{tabularx}{\textwidth}{lYYYYYYYYYYYYY}
\Xhline{2\arrayrulewidth}
\mcs{}                 & \mcs{\textbf{H}}   & \mcs{\textbf{LS}}  & \mcs{\textbf{RS}}  & \mcs{\textbf{LE}} & \mcs{\textbf{RE}}   & \mcs{\textbf{LW}}  & \mcs{\textbf{RW}} & \mcs{\textbf{LH}}   & \mcs{\textbf{RH}}  & \mcs{\textbf{LK}}  & \mcs{\textbf{RK}} & \mcs{\textbf{LA}}   & \mcs{\textbf{RA}}          \\  
\hline
\mcs{{Re-trained OpenPose+PolishNetU}}	&	\mc{98.7}	&	\mc{99.8}	&	\mc{99.7}	&	\mc{99.5}	&	\mc{97.7}	&	\mc{97.0}	&	\mc{97.3}	&	\mc{99.5}	&	\mc{99.5}	&	\mc{98.1}	&	\mc{98.9}	&	\mc{99.2}	&	\mc{98.1}  \\
\mcs{{Re-trained OpenPose}}	            &	\mc{98.8}	&	\mc{99.7}	&	\mc{99.4}	&	\mc{99.3}	&	\mc{97.4}	&	\mc{97.1}	&	\mc{96.0}	&	\mc{99.5}	&	\mc{99.5}	&	\mc{98.2}	&	\mc{98.7}	&	\mc{99.2}	&	\mc{98.4}  \\
\mcs{{Frozen OpenPose+PolishNetU}}   	&	\mc{80.8}	&	\mc{89.6}	&	\mc{44.9}	&	\mc{73.6}	&	\mc{81.6}	&	\mc{79.1}	&	\mc{88.6}	&	\mc{89.3}	&	\mc{85.3}	&	\mc{38.9}	&	\mc{32.8}	&	\mc{36.7}	&	\mc{63.4}  \\
\mcs{{Frozen OpenPose}}             	&	\mc{0.0}	&	\mc{0.0}	&	\mc{0.0}	&	\mc{0.0}	&	\mc{0.0}	&	\mc{0.0}	&	\mc{0.0}	&	\mc{0.0}	&	\mc{0.0}	&	\mc{0.0}	&	\mc{0.0}	&	\mc{0.0}	&	\mc{0.0}   \\
\hline
\mcs{{Re-trained CPN+PolishNetU}}    	&	\mc{92.4}	&	\mc{95.3}	&	\mc{95.2}	&	\mc{94.9}	&	\mc{95.1}	&	\mc{94.9}	&	\mc{94.9}	&	\mc{95.6}	&	\mc{95.6}	&	\mc{95.1}	&	\mc{95.0}	&	\mc{95.3}	&	\mc{95.4}  \\
\mcs{{Re-trained CPN}}	                &	\mc{96.4}	&	\mc{98.5}	&	\mc{98.6}	&	\mc{98.5}	&	\mc{98.4}	&	\mc{97.9}	&	\mc{97.7}	&	\mc{98.6}	&	\mc{98.6}	&	\mc{98.4}	&	\mc{98.4}	&	\mc{98.5}	&	\mc{98.4}  \\
\mcs{{Frozen CPN+PolishNetU}}           &	\mc{48.4}	&	\mc{94.8}	&	\mc{86.9}	&	\mc{52.1}	&	\mc{86.1}	&	\mc{98.1}	&	\mc{100.0}	&	\mc{97.3}	&	\mc{77.4}	&	\mc{79.0}	&	\mc{100.0}	&	\mc{94.4}	&	\mc{85.7}  \\
\mcs{{Frozen CPN}}                      &	\mc{0.0}	&	\mc{0.0}	&	\mc{0.0}	&	\mc{0.5}	&	\mc{0.0}	&	\mc{0.5}	&	\mc{0.0}	&	\mc{0.0}	&	\mc{0.0}	&	\mc{0.0}	&	\mc{0.2}	&	\mc{0.0}	&	\mc{12.8}  \\

\Xhline{2\arrayrulewidth}
\end{tabularx}
\end{center}
\end{table*}

To evaluate the performance of our pipeline on the annotated data, we use the Average Precision (AP), a measure of joint localization accuracy at $5\%$ error margin, similar to the intersection-over-union (IoU) threshold in object detection research \citep{hu2018relation,zhao2019object,zhou2018scale}. First, we sort the predictions by their scores. Then we measure the distances between the predicted and ground-truth keypoints. If this distance is below a threshold, we consider the prediction a true-positive. Finally, we calculate AP by measuring the area under the precision and recall curves. The threshold is defined as a fraction, here $5\%$, of the person's size, where the size is defined as the distance between the person's left shoulder and right hip \citep{andriluka20142d}. In our implementation, the average $5\%$ threshold is equal to approximately $1.2$ pixels or $32$ \textit{mm} considering the size of the input pressure images. Moreover, we also provide another evaluation metric used in pose estimation studies called mean-per-joint-position error (MPJPE) of the predictions \citep{clever20183d, casas2019patient, rhodin2018unsupervised}. MPJPE is measured by averaging the body joint prediction errors in \textit{mm}, calculated in euclidean space.

\begin{figure*}[t]
\begin{center}
{\includegraphics[width=1\textwidth]{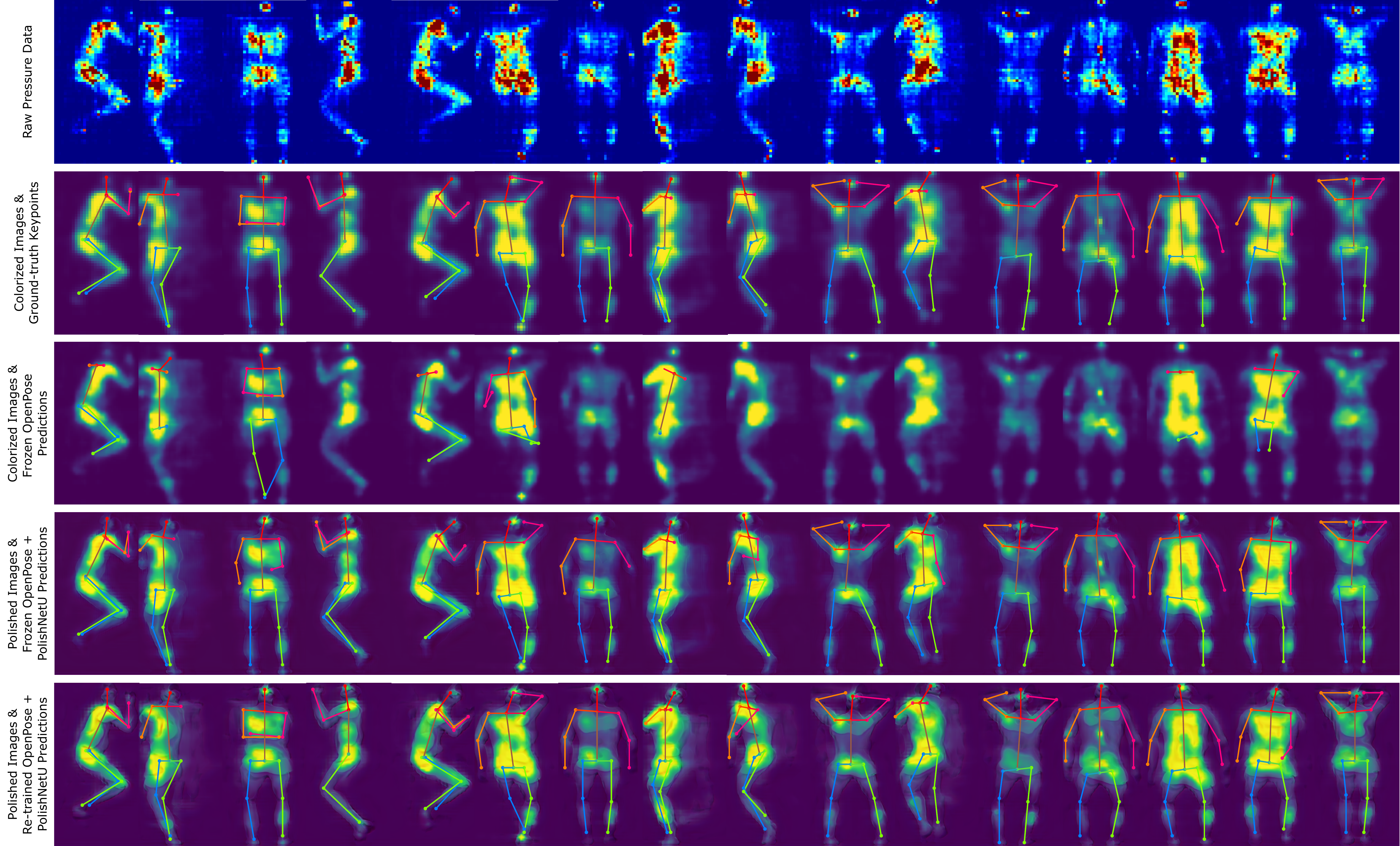}}
\end{center}
   \caption{We provide examples of the performance of the explored architectures for estimating pose from input colorized pressure maps. Here, we show the raw pressure data (first row), colorized pressure data and ground truth keypoints (second row), frozen OpenPose predictions on colorized images (third row), PolishNetU with frozen OpenPose predictions on polished images (fourth row), and PolishNetU with re-trained OpenPose estimations on polished images (fifth row). We removed the predictions that were too far from the actual ground-truth for better presentation. The addition of PolishNetU to the frozen OpenPose causes a boost in pose estimation performance, while re-trained OpenPose with PolishNetU makes slightly more detailed predictions. Furthermore, polished images inherit a better visual fidelity of human body limbs.}
\label{fig:outputs}
\end{figure*}

\begin{table*}
\begin{center}
\caption{The AP5 (\%) - MPJPE (mm) values are presented for two datasets over all possible solutions.}
\label{table:abblation}
\begin{tabularx}{1\textwidth}{YlYYY}
\Xhline{2\arrayrulewidth}
\multirow{2}{*}{\textbf{Pose Estimator}}  & \multirow{2}{*}{\textbf{Model}}  &  \textbf{PolishNetU}               & \multicolumn{2}{c}{\textbf{Dataset}}      \\     \cline{4-5}  
                               &                                  &  \textbf{Fine-tuning}             & \textbf{PmatData}    &  \textbf{HRL-ROS}  \\        
\hline
                                          & Frozen OpenPose                   &       -               & 0.0  - {429.2}         & 0.0  - {279.5}       \\  
OpenPose \citep{cao2016realtime}          & Frozen OpenPose+PolishNetU        &       -               & 78.1 - {29.7}          & 77.4 - {36.6}        \\  
                                          & Re-trained OpenPose               &       -               & 99.8 - {15.1}          & 98.4 - {16.3}        \\ 
\multirow{1}{*}{}                         & Re-trained OpenPose+PolishNetU    &       No              & 99.6 - {15.3}          & 98.6 - {16.2}        \\  
                                          & Re-trained OpenPose+PolishNetU    &       Yes             & 99.9 - {14.8}          & 98.7 - {16.1}        \\ 
\hline
                                          & Frozen CPN                        &       -               & 0.5  - {557.0}         & 0.0  - {208.9}       \\  
CPN \citep{chen2018cascaded}              & Frozen CPN+PolishNetU             &       -               & 71.8 - {78.1}          & 85.6 - {39.2}        \\  
                                          & Re-trained CPN                    &       -               & 99.4 - {20.6}          & 98.0 - {14.4}        \\  
\multirow{1}{*}{}                         & Re-trained CPN+PolishNetU         &       No              & 95.9 - {28.8}          & 95.9 - {18.4}        \\  
                                          & Re-trained CPN+PolishNetU         &       Yes             & 99.6 - {18.4}          & 96.4 - {16.7}        \\  
\Xhline{2\arrayrulewidth}
\end{tabularx}
\end{center}
\end{table*}

\begin{table}
\begin{center}
\caption{The number of parameters used in the models and their backbone feature extractors are presented.} \label{table:params}
\begin{tabularx}{1\columnwidth}{lYY}
\Xhline{2\arrayrulewidth}
\textbf{Model}   & \textbf{Parameters}     &  \textbf{BackBone} \\
\hline
OpenPose     &  52.3M          & Resnet50               \\  
CPN          &  46.0M          & VGG16                  \\  
PolishNetU   &  13.6M          & UNet                   \\  
PolishNet    &  436K           & HourGlass              \\  
\Xhline{2\arrayrulewidth}
\end{tabularx}
\end{center}
\end{table}

We analyze the aforementioned solutions by comparing them based on $AP$, plotted against normalized distances (defined by threshold $\times$ torso length) for different body parts in Figure \ref{fig:PCK_pmat} and \ref{fig:PCK_hrl} for PmatData and HRL-ROS datasets. We omitted the frozen pose estimators without PolishNetU since they showed poor detection rates on both datasets. In contrast, the highest performance in both figures belongs to the models in which the pose estimation network is re-trained, where the combination of PolishNetU and OpenPose is slightly better than the rest, especially for PmatData dataset. Finally, solutions utilizing PolishNetU and frozen pose recognition networks show a weaker performance at $5\%$ threshold compared to the others, but they also reach above $98\%$ detection rate after the $10\%$ threshold ($50mm$) for all of the body parts. It is also shown that the wrists, ankles, and the head are the most challenging body parts, where models containing OpenPose perform better than CPN. A more in-depth comparison of the $AP5$ for our selected models are presented in Tables \ref{table:average_detection_pmat} and \ref{table:average_detection_hrl} for PmatData and HRL-ROS, respectively. We refrained from comparing $AP10$ of our models since in most cases they achieved near perfect accuracy, making the comparisons uninformative. It can be seen that the frozen pose estimators alone are not able to correctly identify poses without PolishNetU.

Several examples depicting the performance of PolishNetU are presented in Figure \ref{fig:outputs}, comparing $3$ of our explored models. Frozen OpenPose is rarely able to predict the correct pose, thus being unreliable. It also miss-identifies the left and right sides of the human body since it was trained on real images that were mostly captured when facing the front of human subjects, as opposed to pressure data that records the image from the backside. As presented in the fourth row, PolishNetU with the frozen OpenPose pipeline has accurately identified the poses for vague input pressure maps while only miss-identifying very blurry areas such as wrists. Finally, the re-trained OpenPose with PolishNetU, illustrated in the fifth row, has made the best pose estimations among other methods, able to identify invisible limbs correctly. 

\begin{figure}[t]
\begin{center}
\includegraphics[width=1\linewidth]{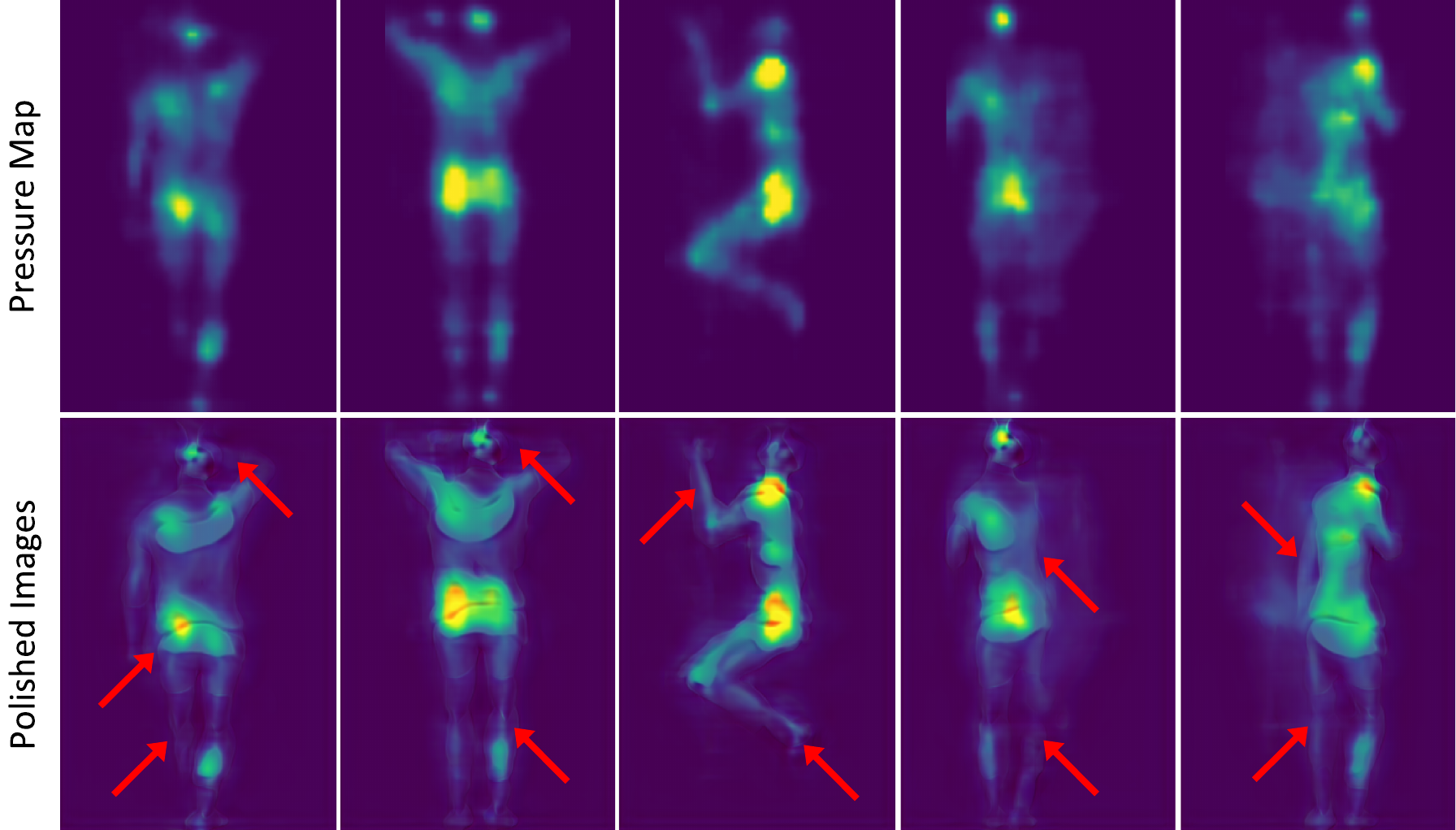}
\end{center}
   \caption{Illustration of examples from PmatData where PolishNetU has reconstructed weak body parts. Specifically note the arms, knees, and the head.}
\label{fig:polished_images}
\end{figure}

Since PolishNetU was trained to synthesize images compatible with the image space by which OpenPose was trained, the polished outputs inherit less noise and show a higher resemblance to common standing human poses from behind. See Figure \ref{fig:polished_images}. Notice that PolishNetU has reconstructed and connected the limbs and weak pressure areas that are not clearly visible in the colorized pressure maps. We have highlighted some of these reconstructed regions in Figure \ref{fig:polished_images}. Moreover, in some instances (i.e. last column of Figure \ref{fig:polished_images}), PolishNetU has even attempted to interestingly synthesize \textit{outfits} for the subjects in order to make the output images look more natural and consistent with the input image space of the pose estimator.

\begin{table*}
\begin{center}
\caption{A comparison of our best results (PolishNetU+Re-trained OpenPose) with other works.} \label{table:comparison}
\begin{tabularx}{1\textwidth}{lYYYYY}
\Xhline{2\arrayrulewidth}
\textbf{Ref}             & \textbf{Dataset}   & \textbf{Evaluation}     &  \textbf{AP5}  &  \textbf{MPJPE {\tiny(\textit{mm})}} &  \textbf{PCK25 {\tiny(\%)}}     \\
\hline
\citep{clever20183d}     &  HRL-ROS           &  Leave $1$ Out          & N/A            &        $73.5^*$              &    N/A                \\  
\citep{davoodnia2019bed} &  PmatData          &  Leave $1$ Out          & N/A            &        N/A                  &    $95.8$           \\  
Ours            &  HRL-ROS           &  Leave $4$ Out          & $98.0$         &        $16.1$               &    $100$            \\  
Ours            &  PmatData          &  Leave $4$ Out          & $99.9$         &        $14.8$               &    $100$            \\  
\Xhline{2\arrayrulewidth}
\multicolumn{6}{l}{\tiny{* denotes the performance of $3$D pose estimation}}
\end{tabularx}
\end{center}
\end{table*}

Finally, we perform an ablation study on the explored models in Table \ref{table:abblation}. We see that the frozen pose estimators without PolishNetU achieve very poor estimation results, where the addition of PolishNetU boosts their performance by a significant amount. Moreover, we observe that re-training the pose estimator block has a higher impact on the performance compared to the previous approach, reducing the MPJPE on both datasets to less than half. Finally, while comparing OpenPose with CPN, we observe that OpenPose outperforms CPN on both datasets based on the $AP5$ metric, which can be because of reasons such as the number of parameters or their features extractor backbones (see Table \ref{table:params}). In contrast, CPN performs better for the HRL-ROS dataset based on the MPJPE criteria.

\begin{figure}[t]
\begin{center}
\includegraphics[width=0.9\linewidth]{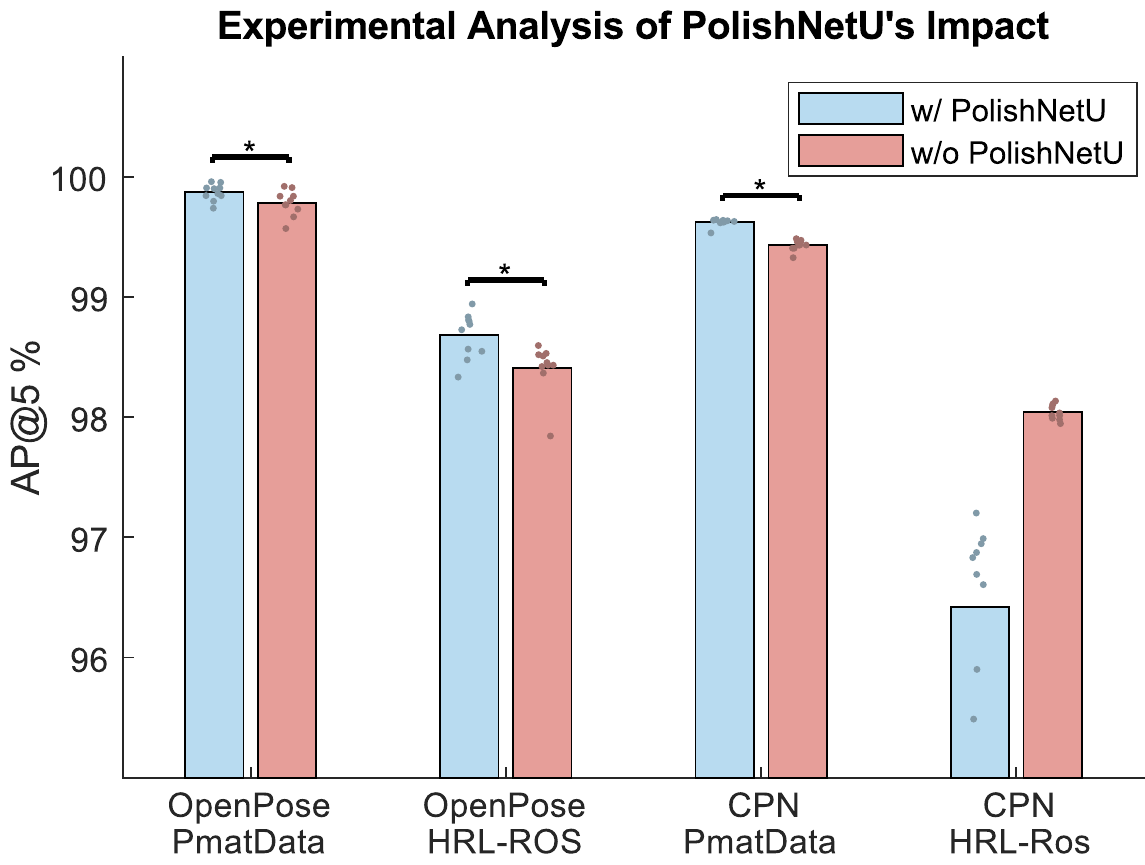}
\end{center}
   \caption{Statistical analysis of $10$ repetitions of our experiments on re-trained pose estimators with and without PolishNetU (fine-tuned) is illustrated, showing that significant improvement is made by PolishNetU in $3$ out of the $4$ scenarios.}
\label{fig:alman}
\end{figure}

The addition of PolishNetU (fine-tuned) to re-trained pose estimators achieves the best results by a small, yet statistically significant margin. We use a non-parametric t-test on $10$ different repetitions of our experiments with random initialization to compare the retrained pose estimators to the retrained pose estimators with PolishNetU (fined-tuned), and achieve $p < 0.05$, showing that the improvements caused by PolishNetU (fine-tuned) are statistically significant. We illustrate this experimental analysis in Figure \ref{fig:alman}, showing the positive impact of PolishNetU in $3$ out of the $4$ cases. Furthermore, Figure \ref{fig:alman} highlights the effectiveness of OpenPose compared to CPN, achieving better performance on both datasets.

\subsection{Discussion and Comparison} \label{discussion}
Although the field of in-bed pose estimation has attracted a considerable amount of recent works, most of the prior works on pressure datasets do not use a unified evaluation method and datasets, making comparisons challenging. We evaluate the solutions based on $2D$ MPJPE and $AP5$, which corresponds to $1.2$ pixels or a $32$ \textit{mm} threshold for correct prediction. We show the effectiveness of PolishNetU by demonstrating the pose estimation performance of pre-existing pose estimators combined with it. Specifically, we show that PolishNetU with a frozen OpenPose achieves near-perfect pose estimation with AP$5$ values of $99.9\%$ and $98.7\%$, and MPJPE of $14.8$ and $16.1$ \textit{mm}, while PolishNetU with a frozen OpenPose performs with AP$5$ of $78.1\%$ and $85.6\%$, and MPJPE of $29.7$ and $39.2$ \textit{mm} on PmatData and HRL-ROS datasets respectively.

Table \ref{table:comparison} compares the results of other works with our best configuration, which is the combination of PolishNetU and the re-trained OpenPose, achieving an $AP5$ of $99.9\%$. Moreover, we show that on the PmatData dataset, we are able to obtain $PCK@25$ of $100\%$, outperforming previous works \citep{davoodnia2019bed}. In another study \citep{clever20183d}, a kinematic-based convolutional neural network was used for $3D$ joint prediction on HRL-ROS dataset, achieving MPJPE of $73.5$ \textit{mm}. Although their model has the advantage of $3D$ joint prediction, several of our explored solutions containing a re-trained pose estimator or the PolishNetU are able to achieve a more accurate prediction in $2D$ space. In a more recent study \citep{clever2020bodies}, the same authors were able to obtain MPJPE of $111.8$ \textit{mm} on a different dataset by training a model called PressureNet on synthetic data and testing it on real images. In another study \citep{casas2019patient}, in-bed pressure data were collected from 6 subjects, reporting an MPJPE of $68$ \textit{mm} using a deep fully convolutional pose estimation model. Lastly, \citep{Chen2018} used a camera-based approach on $3$ subjects. They reported an average accuracy, at $15$ pixels threshold, of $80.5\%$ and $91.6\%$ using a frozen OpenPose and a combination of Kalman filter with a pose estimation network, respectively. Although in some cases our explored methods and datasets are different, we were able to obtain higher performances on a much more larger and complex data space.

% \vspace{-3mm}
\section{Conclusions and Future Work}
In-bed pressure data can provide valuable information for the estimation of a user's pose, which is of high value for clinical and smart home monitoring. However, pressure-based pose estimation deals with a number of challenges, including the lack of large annotated datasets and proper fine-tuned frameworks. Additionally, pressure data impose some inherent limitations such as weak pressure areas caused by supported or raised body parts. In this paper, we explored several end-to-end models for performing pose estimation with in-bed pressure maps, including direct use of off-the-shelf models and re-training them for our purpose. As a part of our analysis, we exploited the novel idea of learning a domain adaptation fully convolutional network, PolishNetU, which generates images as robust representations that work well for common pre-trained pose estimation models, in this case, OpenPose and CPN. This method utilized a compound objective function which integrates the pose identification loss, reconstructing lost body parts caused by weak pressure points, and a pixel loss penalizing large deviations from the original pressure maps. The explored pipelines showed effective performance on highly unclear pressure data. Our evaluation results demonstrated that while re-training the pre-existing pose estimation models have the most impact on performance, if they are kept frozen, PolishNetU can boost the performance significantly as well. Given the performance of PolishNetU with two different pose estimators and on two datasets, we believe this model can be used as a pre-trained block prior to other pose estimation networks for identifying pose from pressure data.

For future work we aim to propose a modified objective function making use of pose priors as a constraint, preventing the model from outputting unlikely poses. Moreover, we aim to investigate the use of generative adversarial networks and integrate a discriminator in our model, which we anticipate, may enhance the reconstruction of weak pressure areas.

\section*{Conflict of Interest}
The authors declare that they have no conflict of interest.

\begin{acknowledgements}
The Titan XP GPU used for this research was donated by the NVIDIA Corporation.
\end{acknowledgements}

\bibliographystyle{spbasic}      % basic style, author-year citations
\bibliography{main_bib}   % name your BibTeX data base

\end{document}